\documentclass[12pt]{l4dc2022}

\newcommand{\R}{\mathbb{R}}

\newcommand{\E}{\mathbb{E}}
\newcommand{\K}{\mathcal{K}}
\newcommand{\KL}{\mathcal{KL}}

\newcommand{\relu}{\text{ReLU}}
\newcommand{\ustress}{u_{\text{stress}}}
\newcommand{\Top}{T_{\text{op}}}
\newcommand{\Tstress}{T_{\text{stress}}}

\newcommand{\rp}[2]{& \multicolumn{1}{c|}{#1} & #2}

\title[ISS Neural ODEs with Applications to Transient Modeling of Circuits]{Input-to-State Stable Neural Ordinary Differential Equations with Applications to Transient Modeling of Circuits}
\usepackage{times}

\author{%
 \Name{Alan Yang} \Email{asyang2@illinois.edu} \\
 \Name{Jie Xiong} \Email{jiex2@illinois.edu} \\
 \Name{Maxim Raginsky} \Email{maxim@illinois.edu} \\
 \Name{Elyse Rosenbaum} \Email{elyse@illinois.edu} \\
 \addr University of Illinois \\ Urbana, IL 61801%
}

\usepackage{graphicx}
\usepackage{pgf,tikz}
\usepackage[arrowmos]{circuitikz}

\begin{document}

\maketitle

\begin{abstract}%
This paper proposes a class of neural ordinary differential equations parametrized by provably input-to-state stable continuous-time recurrent neural networks. The model dynamics are defined by construction to be input-to-state stable (ISS) with respect to an ISS-Lyapunov function that is learned jointly with the dynamics. We use the proposed method to learn cheap-to-simulate behavioral models for electronic circuits that can accurately reproduce the behavior of various digital and analog circuits when simulated by a commercial circuit simulator, even when interconnected with circuit components not encountered during training. We also demonstrate the feasibility of learning ISS-preserving perturbations to the dynamics for modeling degradation effects due to circuit aging.
\end{abstract}


\section{Introduction}

We consider the problem of learning input-to-state stable (ISS) dynamics from observed input and output trajectories. Stability constraints provide an inductive bias that can help a dynamics learning algorithm choose one model over another and may improve the generalization accuracy of the dynamical system model when it is simulated in novel interconnection configurations. In some cases, imposing a stability constraint can guide the learning algorithm and improve performance.

This work was motivated by a need for fast-to-simulate behavioral models of electronic circuits.
A system-on-a-chip (SoC), or larger microelectronic system, is composed of multiple functional blocks, often referred to as \emph{IP blocks}, where IP denotes intellectual property. Prior to manufacturing, simulation is used to verify system performance. Each IP block may be represented by its transistor-level \emph{netlist}, which specifies the interconnection of transistors that comprise its nonlinear dynamics.
Transient simulation of the complete model of the non-linear system can be prohibitively slow, so it is preferable to replace the transistor-level description of each IP block with a cheaper-to-simulate behavioral model.


Transient circuit simulation presents a challenge for dynamics learning. At test time, an adaptive timestep solver is used to simulate the learned dynamics model, which is usually interconnected (in feedback) with external circuits and other IP blocks \citep{Hajj2016}. We refer to these external elements collectively as the \emph{load}. In order to be useful, the model must be accurate when interconnected with a variety of loads, and those loads are generally not known a priori. Due to uncertainty over the load, a model that performs well in training and validation may fail when embedded in a circuit simulation at test time. 


In this work, we assume that the circuit of interest is well-modeled by a system of ODEs. 
A natural approach is to directly learn a parametric system of controlled ODEs
\begin{align}
    \dot x &= f(x,u), \label{eq:node1} \\
    y &= h(x), \label{eq:node2}
\end{align}
which has state $x\in\R^n$, input $u\in\R^m$, and output $y\in\R^p$. $u$ and $y$ consist of node voltages and currents, and possibly their time derivatives. 

In this work, we directly learn a \emph{neural ODE} model of the form \eqref{eq:node1} -- \eqref{eq:node2}, which may be trained by either directly backpropagating through an ODE solver or implicitly differentiating through the solution to the ODEs using an adjoint method \citep{ChenRBD2018}. Our models are trained using interpolated trajectories of $u(t)$ and $y(t)$ obtained from a circuit simulator; this approach is similar to prior works on neural ODE models of physical systems with continuous inputs \citep{KidgerMFL2020, ZhongDC2020a}. 

We focus on the case where $f$ takes the form of a continuous-time recurrent neural network (CTRNN) and $h$ is an affine function of the state. Models of this form are universal approximators on finite time intervals \citep{FunahashiN1993} and, in some cases, infinite time intervals \citep{HansonR2020}. In addition, a CTRNN may be implemented as a generic circuit block using the Verilog-A behavioral modeling language and subsequently simulated by commercial circuit simulators \citep{ChenRR2017}.

In Section \ref{sec:parametrization}, we propose a CTRNN parametrization that is guaranteed to be ISS with respect to a Lur'e-Postnikov type (quadratic plus integral) ISS-Lyapunov function $V(x)$, which has parameters that are learned jointly with the CTRNN model parameters. ISS is a natural \textit{a priori} assumption for many circuits; the state is guaranteed to be bounded given bounded inputs, and the state converges to a unique equilibrium if the input is set to zero.

There has been recent interest in learning neural ODEs jointly with a Lyapunov-like function $V(x)$. For autonomous systems, a sufficient condition for global asymptotic stability (GAS) is to ensure that $V$ is strictly decreasing along any system trajectory, i.e.,
\begin{equation}\label{eq:dissipation_gas}
	\dot V(x)<0 \quad \forall x\ne 0.
\end{equation}
\citet{RichardsBK2018} encourage \eqref{eq:dissipation_gas} via regularization, but do not guarantee that the dissipation inequality holds everywhere. \citet{ManekK2019} and \citet{MassaroliPBPYA2020} define the model dynamics as a function of $V(x)$ such that \eqref{eq:dissipation_gas} holds for all $x$. \citet{CranmerGHBSH2020} and \citet{ZhongDC2020a} considered the related problem of learning dynamics with Lagrangian and Hamiltonian structure, respectively. 

Our approach is similar to that of \citet{ManekK2019} in the sense that we guarantee that a dissipation inequality on $V$ holds everywhere, although we consider ISS, which can be seen as a generalization of GAS to systems with inputs. We use a stability condition that generalizes the ISS condition derived by \citet{Ahn2011}, which is based on a quadratic ISS-Lyapunov function. In another related work, \citet{cao2006state} first learn an unconstrained circuit model and subsequently stabilize the model using nonlinear constrained optimization. In contrast, we build the stability constraint directly into the model parametrization.

Besides providing stability guarantees, we observed that our proposed model parametrization can accelerate training convergence. In this sense, it is related to prior works on regularization methods for accelerating neural ODE training. For example, \citet{FinlayJNO2020} penalized the complexity of the model dynamics while \citet{KellyBJD2020} penalized the forward ODE solution time. Unlike those methods, our stability constraint does not introduce additional penalty terms, which can be difficult to tune.

We also show that our ISS parametrization is directly compatible with \emph{aging-aware} circuit modeling. The dynamics of a circuit drift over time due to semiconductor degradation. Aged dynamics, estimated using physics-based approaches \citep{tu1993berkeley}, can be used to verify lifetime specifications and identify aging-induced failures. \citet{rosenbaum2020machine} directly learn an aging-aware circuit model by choosing the dynamics $f$ in \eqref{eq:node1} and output map $h$ in \eqref{eq:node2} to themselves be learned functions of a periodic \emph{stress waveform} $\ustress$, which is assumed to have been applied to the circuit continuously for an operating time $\Top$ on the order of years. Aging analysis can greatly benefit from fast-to-simulate surrogate models since separate aging simulations are needed to characterize different possible use condition profiles, each of which is specified by a pair ($\ustress$, $\Top$).




Section \ref{sec:parametrization} presents our ISS-constrained model and describes how it can be used for transient circuit simulation, with and without aging effects. Section \ref{sec:experiments} evaluates the proposed methods on a variety of circuit modeling tasks.

\section{Input-to-State Stable Continuous-Time Recurrent Neural Networks}\label{sec:parametrization}

\subsection{Continuous-Time Recurrent Neural Networks}\label{subsec:ctrnns}

We consider controlled neural ODEs of the form
\begin{align}
	\dot x &= -\frac{1}{\tau} x + W\sigma_{\ell}(Ax + Bu + \mu) + \nu, \label{eq:1layer_ctrnn} \\
	     y &= Hx + b, \label{eq:ctrnn_output}
\end{align}
where $x\in\R^{n}$ is the state, $u\in\R^m$ is the input, and $y\in\R^p$ is the output. $\tau>0$ is a positive scalar time constant, and $W,A^\top\in\R^{n\times \ell}$, $B\in\R^{\ell\times m}$, $\mu\in\R^\ell$, and $\nu\in\R^n$ are parameters. The element-wise function $\sigma_{\ell}:\R^{\ell}\to\R^{\ell}$ has the form $\sigma_{\ell}(w) = \begin{bmatrix} \sigma(w_1), \dots, \sigma(w_\ell) \end{bmatrix}^\top$, where $\sigma$ is a strictly increasing, continuous, and subdifferentiable scalar-valued nonlinearity that satisfies $\sigma(0)=0$ and the slope condition 
\begin{equation}\label{eq:slope_condition}
	0 \le \frac{\sigma(r)-\sigma(r')}{r-r'} \le 1, \quad \forall r,r'\in\R, r\ne r'.
\end{equation}
Geometrically, \eqref{eq:slope_condition} means that the graph of $\sigma$ lies within a sector in the first and third quadrants, between the horizontal axis and the line with slope one. 
For example, the conditions on $\sigma(\cdot)$ are satisfied by the rectified linear unit $\relu(\cdot)=\max\{0,\cdot\}$ and the hyperbolic tangent $\tanh(\cdot)$.

The dynamics \eqref{eq:1layer_ctrnn} may be interpreted as a feedforward neural network with a single hidden layer of dimension $\ell$ and a stabilizing term $-x/\tau$, which is similar to ``skip-connections'' in residual networks \citep{HeZRS2016}. We also assume that $\ell\ge n$. Universal approximation results guarantee that a dynamical system with state dimension $n$ can be approximated arbitrarily well by a CTRNN of the form \eqref{eq:1layer_ctrnn} -- \eqref{eq:ctrnn_output}, as long as $\ell$ is sufficiently large \citep{FunahashiN1993,HansonR2020}.

\subsection{Input-to-State Stability}

The notion of input-to-state stability (ISS) was developed as a state-space approach to analyzing the stability of systems with inputs \citep{Sontag2008}. Suppose that \eqref{eq:1layer_ctrnn} has an equilibrium point and, without loss of generality, that the equilibrium is at the origin.
\begin{definition}\label{def:iss}
    The system \eqref{eq:node1} is input-to-state stable (ISS) if there exist a class $\KL$ function\footnote{
    	See \citep{Khalil2002} for definitions of class $\K$, $\K_\infty$, and $\KL$ functions.
    } $\beta$ and class $\K_\infty$ function $\gamma$ such that
	\begin{equation}
		\|x(t)\| \le \beta(\|x_0\|, t) + \gamma(\|u\|_\infty)
	\end{equation}
	for all $t\ge 0$, given any bounded input $u:[0,\infty)\to\R^m$ and initial condition $x(0)=x_0$.
\end{definition}
The ISS property captures the idea that bounded inputs result in bounded state. Moreover, the effect of the initial condition on the trajectory (the transient response) should diminish to zero as $t\to\infty$, with rate bounded by the function $\beta$. 
A sufficient condition for ISS can be found by identifying an appropriate ISS-Lyapunov function $V:\R^n\to\R_+$.

\begin{theorem}{\citep{Khalil2002}}\label{thm:iss_lyapunov}
The system \eqref{eq:node1} is ISS if it admits an ISS-Lyapunov function, i.e., a smooth, positive definite, and radially unbounded
function $V$
for which there exist a positive definite function $\alpha$ and class $\K$ function $g$ such that, for bounded inputs $u$,
    \begin{equation}\label{eq:dissipation_iss}
        \dot V(x,u) := \nabla V(x)^\top f(x,u) \le -\alpha(x)\quad\text{if}\,\, \|x\|\ge g(\|u\|).
    \end{equation}
\end{theorem}
The dissipation inequality \eqref{eq:dissipation_iss} ensures that $V$, and therefore $\|x\|$, cannot grow too large relative to the magnitude of the input. Note that in the absence of inputs, $g(0)=0$, and Theorem \ref{thm:iss_lyapunov} reduces to a sufficient condition for global asymptotic stability. In that case, we refer to the associated function $V$ simply as a Lyapunov function.

\subsection{Lyapunov Diagonal Stability Condition}\label{subsec:lds}

\citet{Forti1995} derived a sufficient condition for which \eqref{eq:1layer_ctrnn} is GAS for constant input $u(t)\equiv u_0$. 

\begin{proposition}\label{prop:forti_tesi}
If the matrix $A$ is full rank, i.e. $\emph{rank}(A)=n$, and there exists a positive diagonal matrix $\Omega=\mathrm{diag}(\omega_1,\dots,\omega_\ell)$ with $\omega_i>0$ for each $i=1,\dots,\ell$ such that
\begin{equation}\label{eq:lds_condition}
	\Omega\Big(AW - \frac{1}{\tau}I\Big) + \Big(W^\top A^\top - \frac{1}{\tau}I\Big)\Omega \prec 0,
\end{equation}
then \eqref{eq:1layer_ctrnn} is GAS for constant input $u(t)\equiv u_0$.
\end{proposition}

If the conditions of Proposition \ref{prop:forti_tesi} hold, we say that the matrix $AW-(1/\tau)I$ is \emph{Lyapunov Diagonally Stable} (LDS). The rank condition on $A$ is not restrictive, since the set of rank-deficient $A$ has measure zero. The proof of Proposition \ref{prop:forti_tesi} makes use of a Lyapunov function of the form
\begin{equation}\label{eq:v_lure}
	V(x) = x^\top P x + 2\sum_{i=1}^\ell \omega_i\int_0^{A_i x} (\sigma_\ell)_i (r)\, \mathrm{d}r,
\end{equation}
where $P\succ 0$ is a positive definite matrix, $A_i$ denotes the $i^{\text{th}}$ row of $A$ in \eqref{eq:1layer_ctrnn} for each $i$, and $\omega_i\ge 0$. If $V$ of the form \eqref{eq:v_lure} can be used to prove 0-GAS for \eqref{eq:1layer_ctrnn}, then it it can also serve as an ISS-Lyapunov function.

\begin{proposition}\label{prop:iss_0gas}
	If the conditions in Proposition \ref{prop:forti_tesi} are satisfied, then \eqref{eq:1layer_ctrnn} is also ISS.
\end{proposition}

The direct extension of Proposition \ref{prop:forti_tesi} to ISS is a consequence of the fact that the LDS condition guarantees that \eqref{eq:1layer_ctrnn} is globally exponentially stable when $u\equiv 0$. In general, 0-GAS is a necessary, but not sufficient, condition for ISS.

\subsection{An Input-to-State Stable Model Parametrization}\label{subsec:iss_new}

Observe that as the matrix $AW$ approaches the zero matrix, the matrix on the left hand side of \eqref{eq:lds_condition} approaches $-\frac{2}{\tau}\Omega$, which is negative definite. Therefore, we may stabilize a given CTRNN by scaling $AW$. Here, we consider a parametrized matrix $A_\theta\in\R^{\ell\times n}$, and define $A$ to be
\begin{equation}\label{eq:parametrization_A}
	A = \frac{1}{\rho(\tau,A_\theta,W,\Omega)+1}A_\theta,
\end{equation}

In the following, let $\lambda_{\max}(M)$ denote the largest eigenvalue of a symmetric matrix $M$. 
\begin{theorem}\label{thm:stable_ctrnn}
For any $\delta>0$, let
\begin{equation}\label{eq:parametrization_rho}
	\rho(\tau,A_\theta,W,\Omega) = \emph{\relu}\bigg(\frac{\tau}{2}\lambda_{\max}\big(\Omega^{1/2} A_\theta W\Omega^{-1/2} + \Omega^{-1/2}W^\top A_\theta^\top\Omega^{1/2}\big) - 1 + \delta\bigg),
\end{equation}
Then, \eqref{eq:1layer_ctrnn} with $A$ given by \eqref{eq:parametrization_A} is ISS. 
\end{theorem}

Using \eqref{eq:parametrization_rho}, the scalar $\rho$ is made large enough to ensure that the LDS condition holds, if the condition does not already hold with $A=A_\theta$. The hyperparameter $\delta$ controls the minimum dissipation rate of $V$; smaller values of $\delta$ allow for longer transients. The model may be directly trained using a gradient descent method, since the stability constraint is built into the definition of $A$. $\Omega$ parametrizes $V$ in \eqref{eq:v_lure}, and may be either learned with the rest of the model parameters or fixed, e.g. to $\Omega=I$.

%


\subsection{Training CTRNN Neural ODEs}\label{subsec:training}

At the start of a transient simulation, a circuit simulator sets the initial condition of its state variables to an equilibrium point given the initial input. For the model \eqref{eq:node1} -- \eqref{eq:node2}, the circuit simulator sets $x(0)=x_0$, where $x_0$ satisfies
\begin{equation}\label{eq:rootfind}
	0 = f(x_0,u(0)).
\end{equation}
This is done using a numerical root-finding scheme, such as the Newton-Raphson method. The equilibrium condition \eqref{eq:rootfind} does not necessarily uniquely define $x_0$, and a user-specified initial condition may need to be provided. Fortunately, if $f$ is given by a CTRNN \eqref{eq:1layer_ctrnn} satisfying the LDS condition, then Proposition \ref{prop:forti_tesi} guarantees the existence of a unique $x_0$ that satisfies \eqref{eq:rootfind} for any $u(0)$. During model training, we set the initial condition by numerically solving \eqref{eq:rootfind}. In order to obtain the necessary derivatives for optimization, we implicitly differentiate through the root-finding operation \citep{BaiKK2019}.

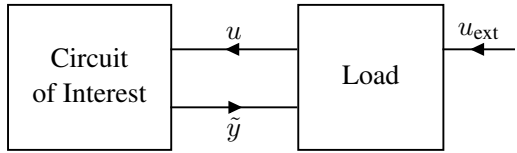
\begin{figure}[htbp]
\centering
\floatconts
  {fig:interconnection}
  {\caption{Simulation diagram for an interconnection between a circuit of interest and a load.}}
  {%
\resizebox{0.45\textwidth}{!}{%
\begin{circuitikz}[line width=0.75pt, american]
    \node[draw, minimum width=2cm, minimum height=2cm, anchor=south west] at (1.5,-.5) {\begin{tabular}{c} Circuit \\ of Interest \end{tabular}};
    \node[draw, minimum width=2cm, minimum height=2cm, anchor=south west] at (5.5,-.5) {\begin{tabular}{c} Load \end{tabular}};
    \draw (8.5,0.9) to[short, i_=$u_{\text{ext}}$] (7.5,0.9);
    \draw (3.75,0.9) to[short, i^<=$u$] (5.5,0.9);
    \draw (5.5,0.1) to[short, i^<=$\tilde y$] (3.75,0.1);
\end{circuitikz}
    }
}
\end{figure}

In our experiments, we consider the simulation setup illustrated in Figure \ref{fig:interconnection}. In a simulation, the circuit of interest is interconnected with a load, which itself may be driven by an external signal $u_{\text{ext}}$. We assume that both the load and $u_{\text{ext}}$ are random and have known distributions. We train our models using $N$ input and output trajectories $u^{(i)}$ and $\tilde y^{(i)}$ for $i=1,\dots,N$. Each pair ($u^{(i)},\tilde y^{(i)})$ is obtained by simulating the system in Figure \ref{fig:interconnection} on a time interval $[0,T]$ with initial condition defined by \eqref{eq:rootfind}, using a random instantiation of the load and $u_{\text{ext}}$. Like \citet{KidgerMFL2020}, we obtain continuous trajectories $u^{(i)}$ and $\tilde y^{(i)}$ by interpolating the solution points provided by the circuit simulator.

The model parameter learning problem is given by the optimization problem
\begin{equation}\label{eq:loss}
\text{minimize}\,\, \frac{1}{N}\sum_{i=1}^N \frac{1}{T}\int_0^T \big(\tilde y^{(i)}(t) - y^{(i)}(t)\big)^2 \, \mathrm{d}t,
\end{equation}
where $y^{(i)}$ is the predicted output. We estimate the integral in \eqref{eq:loss} by a Monte Carlo estimate as follows. Let $S$ be a random variable uniformly distributed on $[0,T]$. Then, we have $\frac{1}{T}\int_0^T \big(\tilde y^{(i)}(t) - y^{(i)}(t)\big)^2 \, dt = \E[\big(\tilde y^{(i)}(S) - y^{(i)}(S)\big)^2]$ for each $i$, and so we may estimate the expectation using $\E[\big(\tilde y^{(i)}(S) - y^{(i)}(S)\big)^2] \approx \frac{1}{K}\sum_{j=1}^K \big(\tilde y^{(i)}(S_j) - y^{(i)}(S_j)\big)^2$, where $S_1,\ldots,S_K$ are i.i.d.\ copies of $S$.

\subsection{Aging-Aware Neural ODEs}\label{subsec:aging}

Conventionally, circuit aging simulation involves two transient simulations of the complete transistor-level netlist \citep{tu1993berkeley}. In the first step, the fresh circuit ($\Top=0$) is simulated subject to  $\ustress$ on a short time horizon $\Tstress\ll\Top$\footnote{Modern circuits have nanosecond-scale signal periods; $\Tstress$ on the order of tens of nanoseconds is usually sufficient.} 
to estimate the per-transistor stress profile. Each transistor's dynamics is subsequently age-adjusted assuming that the $\Tstress$-periodic input $\ustress$ is applied for time $\Top$, which is typically on the order of years. In the second step, the circuit is re-simulated using the age-adjusted transistor models, subject to a possibly new input $u$.

\citet{rosenbaum2020machine} proposed a two-step learning approach to learn aging-aware models; we extend that approach to include the ISS constraint. In the first step, we learn a ``fresh'' CTRNN 
\begin{align*}
    \dot x &= -\frac{1}{\tau_0} x + W_0\sigma_\ell(A_0x + B_0u + \mu_0) + \nu_0, \\
    y &= H_0x + b_0,
\end{align*}
corresponding to $\Top=0$ using the approach in Subsection \ref{subsec:training}. In the second step, we fix $\tau_0,W_0,\dots,b_0$ and form an aging-aware CTRNN \eqref{eq:1layer_ctrnn} -- \eqref{eq:ctrnn_output} whose parameters are given by 
\begin{align*}
    \tau &= \tau_0 + \Delta_\tau(\ustress,\Top), \\
    W &= W_0 + \Delta_W(\ustress,\Top), \\
    &\dots \\
    b &= b_0 + \Delta_b(\ustress,\Top),
\end{align*}
where $\Delta_\tau,\Delta_W,\dots,\Delta_b$ are learned parameter perturbation functions. The model structure is suitable because, for realistic use conditions, the stress-induced drift in the dynamics will be relatively small, as illustrated by the example in Figure \hyperref[subfig:predictedinv]{2\textit{(d)}}. An ISS aging-aware model may be obtained by setting $A_\theta = A_0 + \Delta_A(\ustress,\Top)$ in \eqref{eq:parametrization_A}. The perturbation functions are learned using randomly-generated $\ustress$, $\Top$, and corresponding output trajectories $\tilde y$ obtained from the circuit simulator.

\section{Experiments}\label{sec:experiments}

\subsection{Test Cases}

\paragraph{Common Source Amplifier.} This circuit is a one-transistor, two-port voltage amplifier connected to resistor-capacitor (RC) loads with randomly generated values. The system input $u_{\text{ext}}$ is driven by a random piecewise linear voltage source. The goal is to predict the currents at the input and output ports, given the port voltages. Our CTRNN models had dimensions $n=6$, $\ell=14$, and $m=p=2$.

\paragraph{Continuous-Time Linear Equalizer (CTLE).} The CTLE is a five-transistor differential amplifier with two input ports and two output ports. It is designed to compensate for signal distortion that occurs when digital data are transmitted between two chips. The system input $u_{\text{ext}}$ is given by the output of a pseudorandom bit sequence generator passed through a USB serial link, and the ports are connected to randomly-generated RC loads. The goal is to predict the input port currents and output port voltages, given the input port voltages, their time derivatives, and the output port currents. Our CTRNN models had dimensions $n=20$, $\ell=30$, $m=6$, and $p=4$.

\paragraph{Large IP Block.} 
This test circuit contains between 1000 and 2000 transistors; the exact number is unknown because the circuit is described by an encrypted netlist. Unlike the previous two test cases, this circuit is used with a known, fixed load. The goal is to predict two output voltages given seven input voltages. The inputs are driven by $u_{\text{ext}}$ given by the outputs of seven pseudorandom bit sequence generators. Our CTRNN models had dimensions $n=20$, $\ell=30$, $m=7$, and $p=2$. 

\begin{figure}[hbp]
\centering
\floatconts
    {fig:predicted_waveforms} 
    {\caption{Examples of the proposed ISS CTRNN predictions when simulated by a circuit simulator.}} 
    {%
      \subfigure[Amplifier]{%
        \label{subfig:predictedamp}%
        \includegraphics[width=0.48\textwidth]{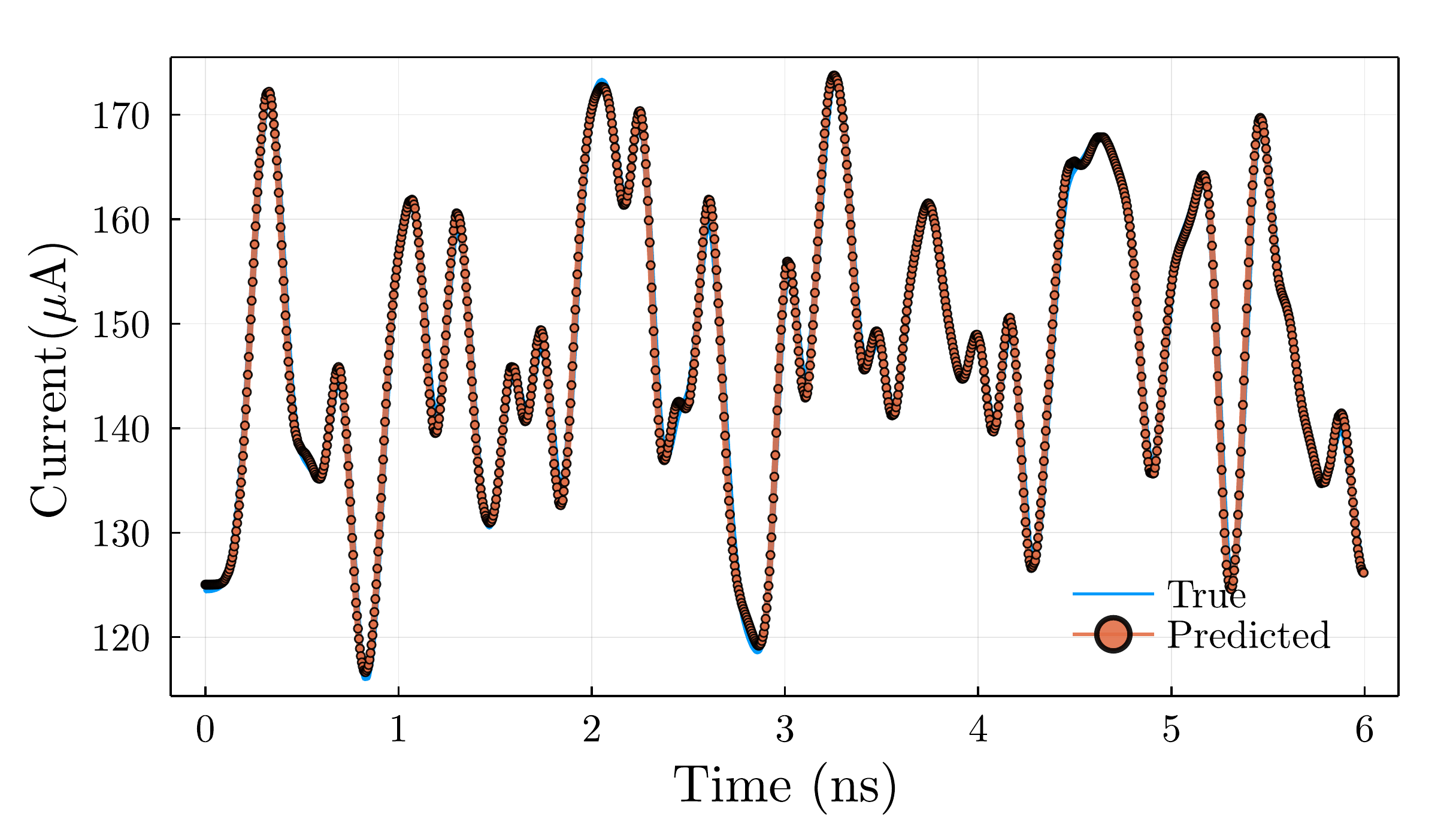}
      }
      \subfigure[CTLE]{%
        \label{subfig:predictedctle}%
        \includegraphics[width=0.48\textwidth]{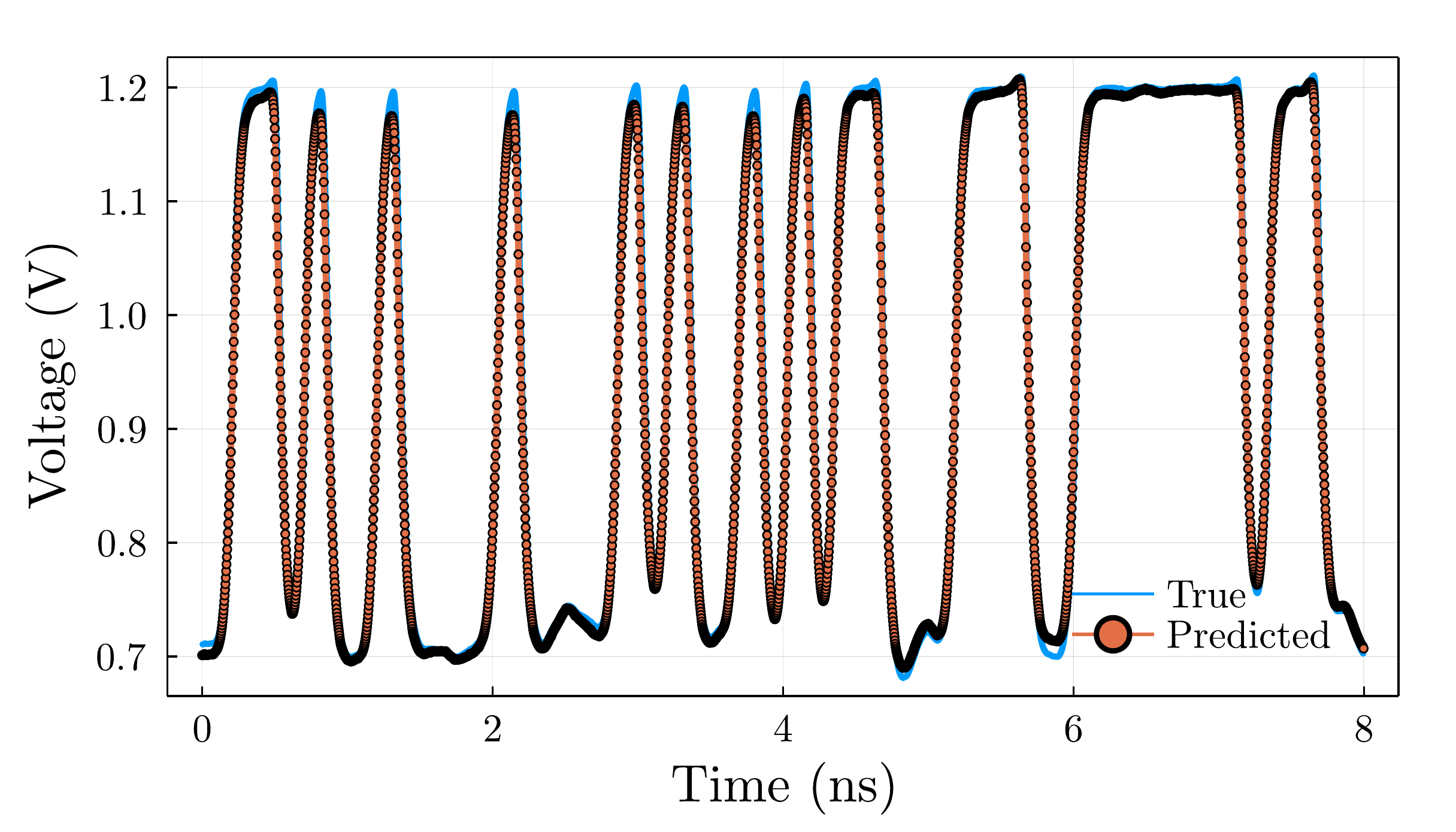}
      }
      \subfigure[IP block]{%
        \label{subfig:predictedadi}%
        \includegraphics[width=0.48\textwidth]{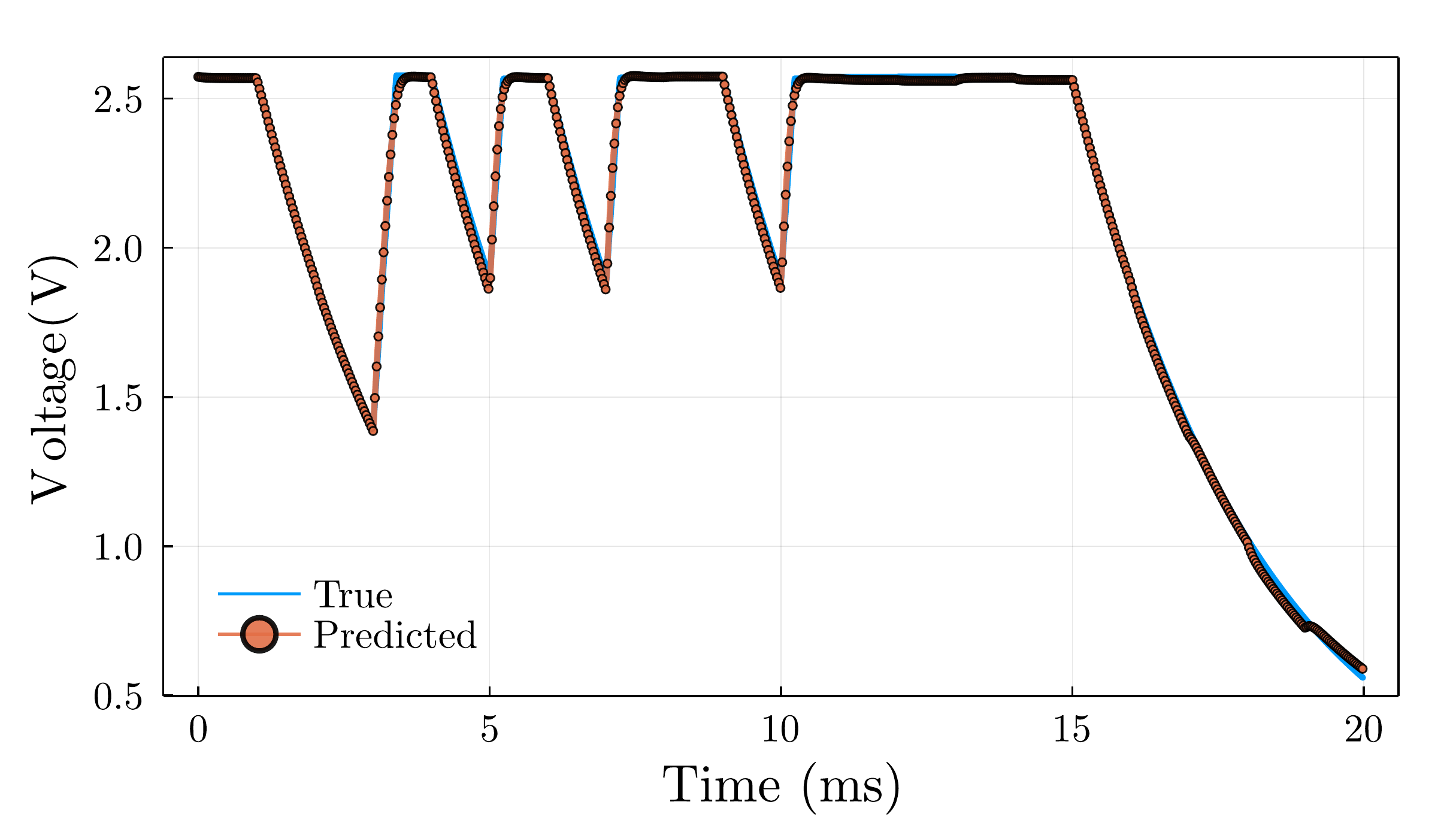}
      }
      \subfigure[Inverter Chain]{%
        \label{subfig:predictedinv}%
        \includegraphics[width=0.48\textwidth]{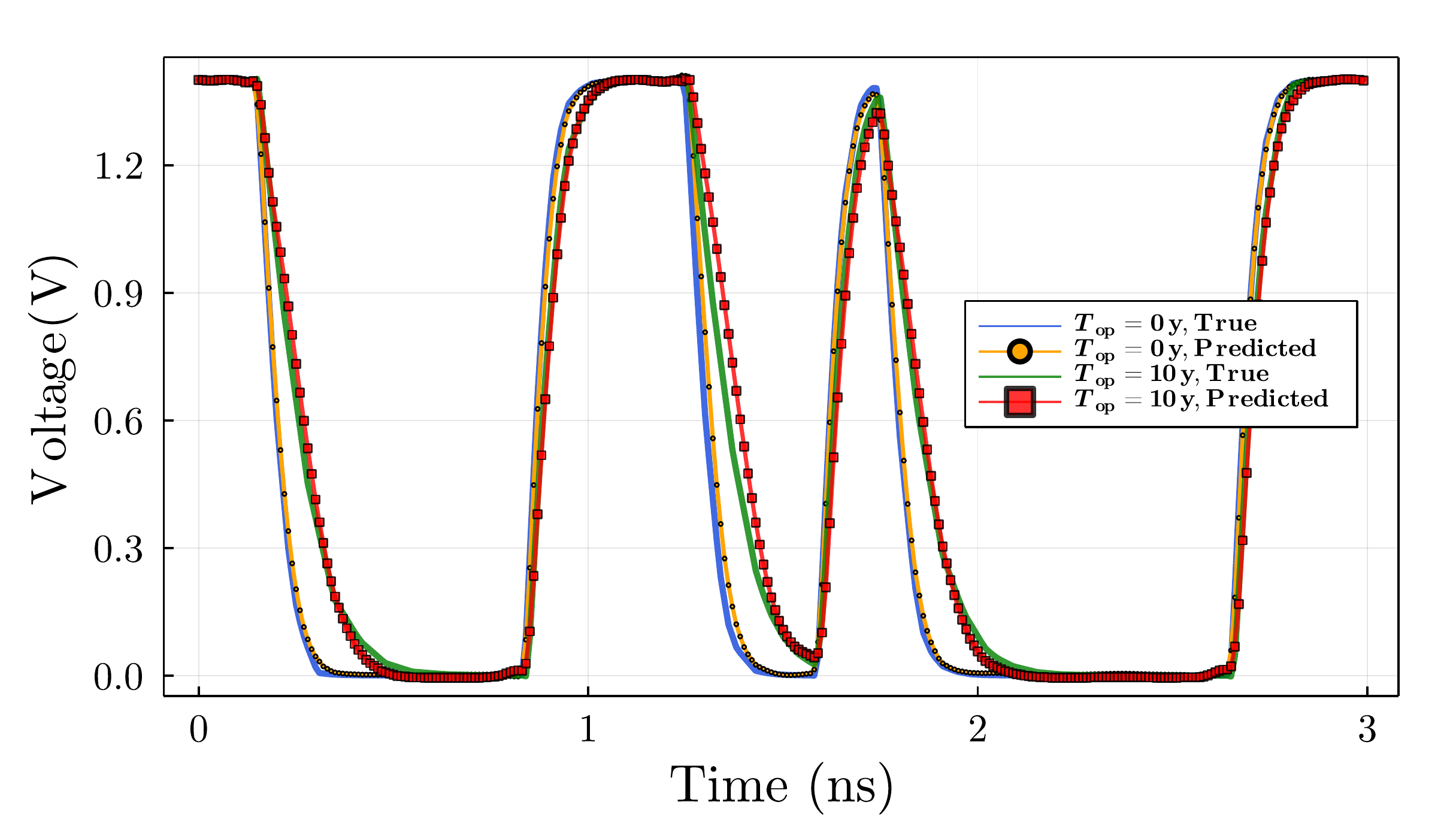}
      }
    }
\end{figure}

\paragraph{Inverter Chain with Aging Effects.} The last test circuit is a chain of nine cascaded digital inverters; this circuit is often used to benchmark aging. For this test case, aging-induced degradation slows down the dynamics; Figure \hyperref[subfig:predictedinv]{2\textit{(d)}} illustrates the delay between the outputs of a fresh and aged circuit. The model outputs two port currents given two port voltages, given the stress profile $(\ustress,\Top)$. The system input $u_{\text{ext}}$ was driven by a random piece-wise linear voltage source, and random capacitive loads were connected to the output port. Aging analysis was performed using random piece-wise linear $\ustress$ and random $\Top$ sampled from a log uniform distribution from 0.001 to 10 years. We considered a special case of \eqref{eq:1layer_ctrnn} with $W=I$ and $\nu=0$ with dimensions $n=\ell=20$ and $m=p=2$, and learned parameter perturbations only for $A$, $B$, and $\mu$. $\Delta_A$, $\Delta_B$, and $\Delta_\mu$ were implemented by single-layer gated recurrent unit (GRU) network \citep{cho2014learning} with hidden state dimension 20.

\subsection{Results}

We trained the CTRNNs by directly backpropagating through the order three Bogacki-Shampine ODE solver
with the ADAM optimizer \citep{KingmaB2015}. We took $\sigma_\ell$ to be $\relu$ with a bias term, and we used $\delta=10^{-3}$ in the stability constraint \eqref{eq:parametrization_rho}. Model parameters were initialized randomly, with the constraint that \eqref{eq:lds_condition} held with $\Omega=I$.  Each dimension of the inputs and outputs in the training data was separately normalized to $[-1,1]$ prior to training, and the time horizon $T$ was scaled up to be on the order of seconds (instead of nanoseconds). The models used in the first three test cases were trained using the Julia package DiffEqFlux \citep{RackauckasMMCZSSRE2020}; the aging-aware models were trained using the Python package torchdiffeq \citep{ChenRBD2018}. Learned models were implemented in Verilog-A and simulated using the Spectre circuit simulator \citep{Cadence2020}.

Table \ref{table:performance} compares three different training methods: CTRNN with no constraints (Baseline), the stability constraint \eqref{eq:parametrization_A} (Proposed), and the stability constraint with $\Omega$ fixed to be the identity matrix (Proposed, $\Omega=I$). The table shows the mean squared error (MSE) of the predicted model outputs measured on a held-out validation set of input and output waveforms (``Valid.'') and measured when the model is simulated by the circuit simulator as a Verilog-A model (``Test''). The ``Test'' MSE values are averaged across 100 simulations, each with random instantiations of load, $u_{\text{ext}}$, and stress profile ($\ustress,\Top$), in the aging-aware inverter chain test case. The MSE for both ``Valid.'' and ``Test'' are computed after applying the aforementioned normalization to $[-1,1]$. Figure \ref{fig:predicted_waveforms} shows example simulations of the proposed ISS CTRNN models, carried out by Spectre.


\begin{table}[htbp]
\centering
\tableconts
{table:performance} 
{\caption{MSE metrics. The reported MSE have been multiplied by a factor of 1000.}} 
{%
\begin{tabular}{|l||cc||cc||cc||cc|}
\hline
 & \multicolumn{2}{c||}{Amplifier}     & \multicolumn{2}{c||}{CTLE}          & \multicolumn{2}{c||}{IP Block}     & \multicolumn{2}{c|}{Inverter Chain}          \\ 
 \hline
Model Type & \multicolumn{1}{c|}{Valid.} & Test & \multicolumn{1}{c|}{Valid.} & Test & \multicolumn{1}{c|}{Valid.} & Test & \multicolumn{1}{c|}{Valid.} & Test  \\ \hline\hline
Proposed                \rp{0.234}{0.263}   \rp{0.805}{\bf 0.939}    \rp{\bf 0.031}{\bf 0.250}   \rp{0.280}{\bf 0.918}   \\ \hline
Proposed ($\Omega=I$)   \rp{0.260}{0.314}   \rp{1.03}{4.44}    \rp{0.142}{0.2943}   \rp{\bf 0.232}{1.93}    \\ \hline
Baseline                \rp{0.243}{0.279}   \rp{0.714}{3.1447}    \rp{2.93}{15.9}     \rp{0.238}{3.50}        \\ \hline
\end{tabular}
}
\end{table}

Across all test cases, the errors on the held out test set were lower than the errors accumulated when tested in the circuit simulator. This is expected, since the model is simulated in open-loop with the validation set data rather than in closed-loop with the circuit simulator. The proposed constraints uniformly outperformed the baseline learning method, and the stability constraint with learned $\Omega$ almost always outperformed the stability constraint with $\Omega$ fixed to the identity.

For the IP block test case, the ISS constraint was necessary for learning an accurate model. Without the stability constraint, the training struggled to escape a local minimum. This can be seen in the validation losses over the course of training, shown for the IP block and CTLE test cases in Figure \ref{fig:valid_losses}. Although the effect was less pronounced, the ISS constraints were able to stabilize training for the CTLE test case as well.

\section{Conclusion}

In this work, we proposed provably ISS CTRNN neural ODE models for transient circuit simulation. The CTRNN models are constructed in terms of an ISS-Lyapunov function $V$ such that a dissipation inequality on $V$ is satisfied, and can yield better models than baseline unconstrained training approaches. In principle, this type of approach may be extended to impose other dissipation or invariance conditions on $V$, for example conditions based on passivity or energy conservation; that is a suitable subject for future work. 

\begin{figure}[hbp]
\centering
\floatconts
  {fig:valid_losses}
  {\caption{Comparison of validation losses over the course of training.}}
  {%
      \subfigure[IP Block]{%
        \label{subfig:predicted_amp}
        \includegraphics[width=0.45\textwidth]{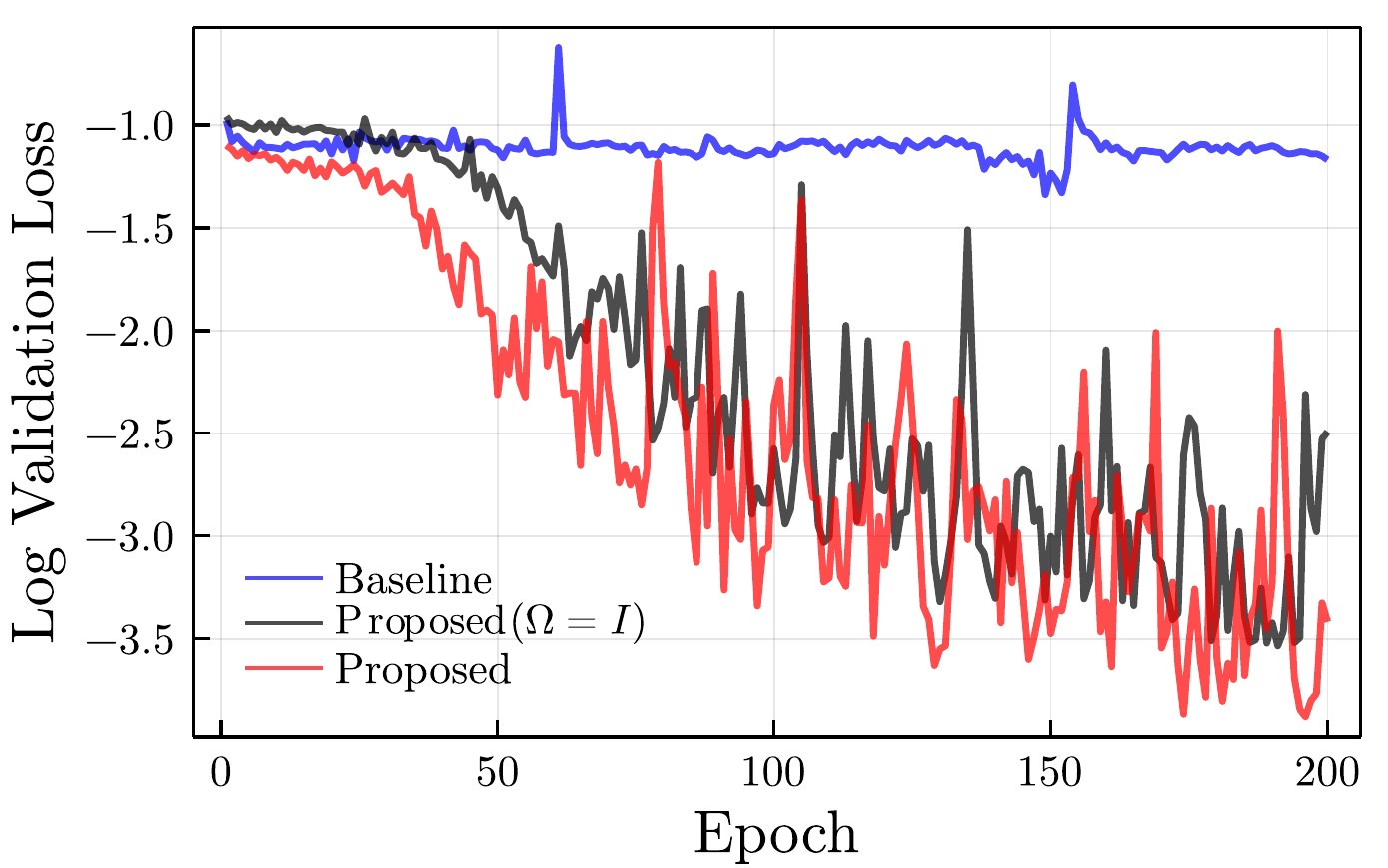}
      }
      \subfigure[CTLE]{%
        \label{subfig:predicted_ctle}
        \includegraphics[width=0.45\textwidth]{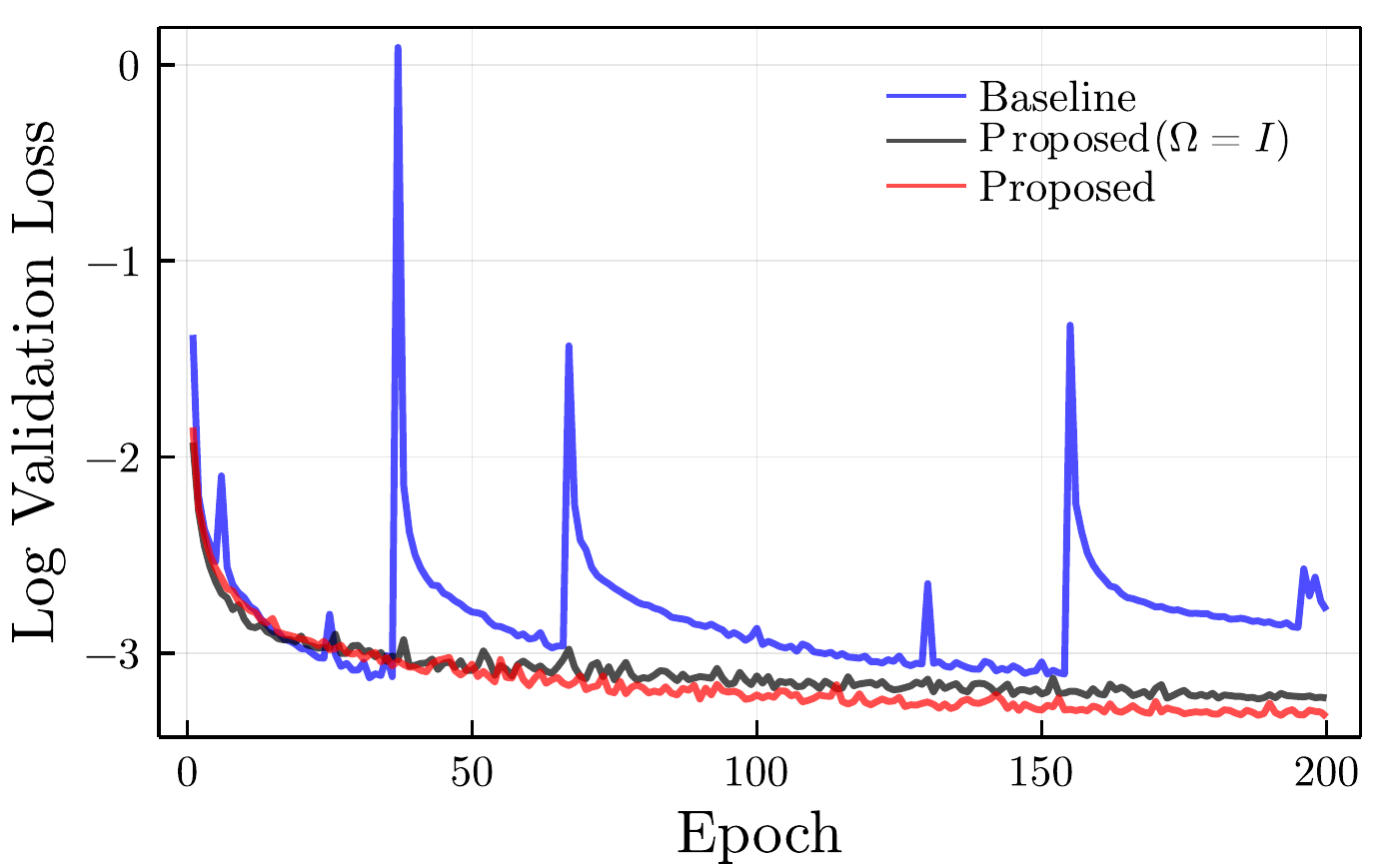}
      }
  }
\end{figure}

\acks{This work was funded in part by the NSF under CNS 16-24811 and the industry members of the CAEML I/UCRC, and in part by the Illinois Institute for Data Science and Dynamical Systems (iDS${}^2$), an NSF HDR TRIPODS institute, under award CCF-1934986.}

\appendix
\section{Omitted Proofs}

\subsection{Proof of Proposition \ref{prop:forti_tesi}}
    Consider any constant input $u(t)=u_0$. The change of coordinates $z=Ax-Bu_0-\mu$ gives
	\begin{equation}\label{eq:iss_transformed_ctrnn}
		\dot z = -\frac{1}{\tau}z + AW\sigma_\ell(z) + A\nu  - \frac 1 \tau Bu_0 - \frac 1 \tau \mu.
	\end{equation}
	Theorem 4 of \citet{Forti1995} guarantees that \eqref{eq:iss_transformed_ctrnn} has a unique equilibrium point $z_0$ which is GAS with respect to a Lyapunov function $V(z)$, where $V$ is of the form \eqref{eq:v_lure}. Since $\ell\ge n$ and $A$ is full rank, $x_0 = (A^\top A)^{-1}A^\top z_0$ is the unique equilibrium point of \eqref{eq:1layer_ctrnn}, and is GAS with respect to $V((A^\top A)^{-1}A^\top z)$, which can also be written in the form \eqref{eq:v_lure}.

\subsection{Proof of Proposition \ref{prop:iss_0gas}}
    Suppose that \eqref{eq:1layer_ctrnn} is 0-GAS with respect to the Lyapunov function \eqref{eq:v_lure}. With $u\equiv 0$, the time derivative of $V$ along trajectories of $x$ is given by
	\begin{align}
		\dot V(x) 
		&= 2(x^\top P + \sigma_\ell(Ax)^\top \Omega A)(-x/\tau + W\sigma_\ell(Ax)) \nonumber \\
		&= \begin{bmatrix}
			x \\ \sigma_\ell(Ax)
		\end{bmatrix}^\top
		\begin{bmatrix}
			-2P/\tau & PW - \frac{1}{\tau}A^\top \Omega  \\
			W^\top P - \frac{1}{\tau}\Omega A & \Omega AW + W^\top A^\top \Omega
		\end{bmatrix}
		\begin{bmatrix}
			x \\ \sigma_\ell(Ax)
		\end{bmatrix}. \label{eq:Vdot_gas}
	\end{align}
    Since the origin is a GAS equilibrium point, $\dot V(x)<0$ for all $x\ne 0$, which implies that \eqref{eq:Vdot_gas} is a negative definite quadratic form, i.e., there exists a $\lambda>0$ such that $\dot V(x)\le -\lambda (\|x\|^2 + \|\sigma_\ell(Ax)\|^2) \le -\lambda\|x\|^2$. Due to the slope condition \eqref{eq:slope_condition}, $V(x)$ has a quadratic upper bound, and so by Theorem 4.10 of \citet{Khalil2002}, the unforced system is globally exponentially stable at the origin. Finally, Lemma 4.6 in \citet{Khalil2002} gives ISS.

\subsection{Proof of Theorem \ref{thm:stable_ctrnn}}

To simplify the notation, we write $\rho(\tau,A_\theta,W,\Omega)$ as $\rho$ with the arguments omitted. Since $\relu(\cdot) = \max\{0,\cdot\}$, we have $\rho \ge  \frac{\tau}{2}\lambda_{\max}\big(\Omega^{\frac{1}{2}} A_\theta W\Omega^{-\frac{1}{2}} + \Omega^{-\frac{1}{2}}W^\top A_\theta^\top\Omega^{\frac{1}{2}}\big) - 1 + \delta$.
Dividing both sides by $\rho+1$ and rearranging gives $1\ge \frac{\tau}{2}\lambda_{\max}\big(\Omega^{\frac{1}{2}} AW\Omega^{-\frac{1}{2}} + \Omega^{-\frac{1}{2}}W^\top A^\top\Omega^{\frac{1}{2}}\big) + \frac{\delta}{\rho+1}$, where $A=\frac{A_\theta}{\rho+1}$.
Since $\lambda_{\max}(M)I\succeq M$ for symmetric $M$, $I \succeq \frac{\tau}{2}\big(\Omega^{\frac{1}{2}} AW\Omega^{-\frac{1}{2}} + \Omega^{-\frac{1}{2}}W^\top A^\top\Omega^{\frac{1}{2}}\big) + \frac{\delta}{\rho+1}I$. Finally, multiplying by $\Omega^{\frac 1 2}$ on the left and right sides and rearranging shows the LDS condition
\begin{equation*}
    	\Omega \big(AW - \frac{1}{\tau}I\big) + \big(W^\top A^\top - \frac{1}{\tau}I\big)\Omega \preceq - \frac{2\delta}{\tau(\rho+1)}\Omega \prec 0.
\end{equation*}


\bibliography{l4dc2022-sample}
\end{document}